\documentclass{svproc}
\usepackage{url}

\usepackage{amssymb,amsfonts,amsmath}
\usepackage{array}
\usepackage{graphicx} 
\usepackage{epstopdf}
\usepackage{bmpsize}
\usepackage{multirow,comment}
\usepackage{multicol}
\usepackage{footmisc}
\usepackage[caption=false]{subfig}
\usepackage{xcolor}
\usepackage{float}
\usepackage{cite}

\newcommand{\bC}{{\mathbb C}}
\DeclareMathOperator{\Var}{{Var}}

\begin{document}
\setlength{\abovedisplayskip}{5pt}
\setlength{\belowdisplayskip}{5pt}
\setlength{\abovedisplayshortskip}{2pt}
\setlength{\belowdisplayshortskip}{2pt}
\mainmatter              % start of a contribution
\title{Using monodromy to statistically estimate the number of solutions}
\titlerunning{Statistical estimation using monodromy}  % abbreviated title (for running head)
\author{Jonathan D. Hauenstein \and Samantha N. Sherman
}
\authorrunning{J.D. Hauenstein and S.N. Sherman} % abbreviated author list (for running head)
\institute{University of Notre Dame, Notre Dame, IN 46617, USA\\
\email{\{hauenstein,ssherma1\}@nd.edu}
}

\maketitle              % typeset the title of the contribution

\begin{abstract}
Synthesis problems for linkages in kinematics
often yield large structured 
parameterized polynomial systems 
which generically have far fewer solutions than traditional upper bounds would suggest.  This paper describes statistical models for estimating the
generic number of solutions of such 
parameterized polynomial systems.  The new approach extends previous work on success ratios of parameter homotopies to using monodromy loops as well as the addition of a trace test that provides 
a stopping criterion for validating that all solutions have been found. Several examples are presented demonstrating the method including Watt I six-bar motion generation problems.

\end{abstract}
%
% has own theorem, prop, corr, proof environments.
% also recommends using eqnarray environment, i.e.,
%\begin{eqnarray*}
%  \dot{x}&=&JH' (t,x)\\
%  x(0) &=& x(T)
%\end{eqnarray*}

\section{Introduction}\label{Sec:Introduction}

\begin{figure}[!htbp]
\vspace{-0.5cm}
\begin{center}
%\begin{tabular}{ccc}
\subfloat[Four Bar]{
 \includegraphics[scale=0.45]{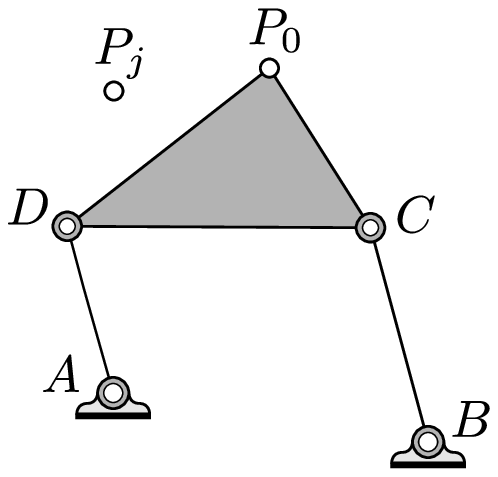}\label{fig:4BarLabeled}}
~ \hspace{1in}
\subfloat[Watt 1]{
\includegraphics[scale=0.4]{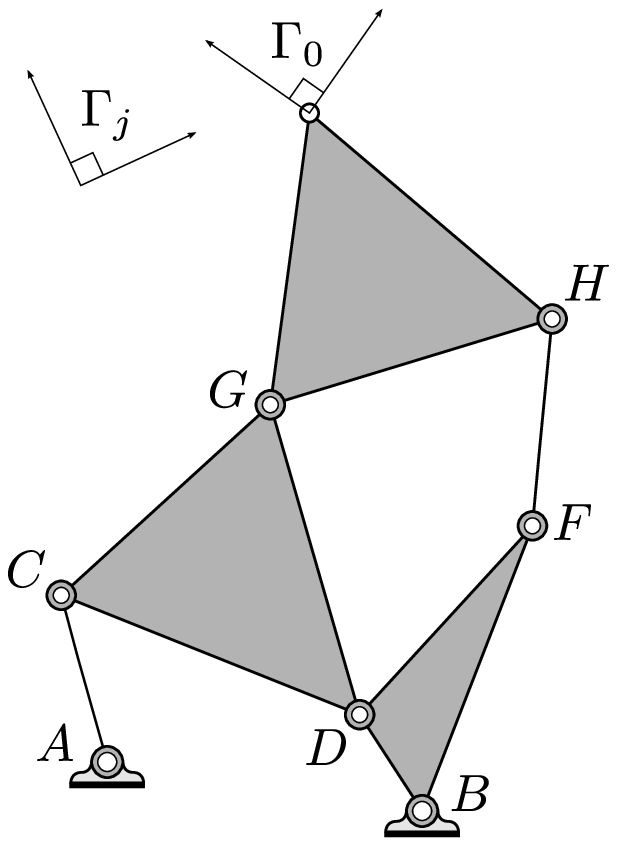}\label{fig:W1Labeled}}
%\end{tabular}
\caption{Four-bar mechanism and Watt I six-bar mechanism showing initial point and task position,
respectively, and one offset.}
\label{fig:W1and4BarLabeled}
\end{center}
\vspace{-0.6cm}
\end{figure}

In linkage design, synthesizing rigid body 
linkages yield
polynomial systems~\cite{AlgKinematics} which are parameterized
by the desired tasks.  
For example, 
Alt \cite{Alt} considered synthesizing 
four-bar linkages specifying 9 path points 
as pictorially represented in Figure~\ref{fig:W1and4BarLabeled}(a).
Once a synthesis problem is formulated,
a natural first step is to estimate 
the number of solutions to decide if
it is practical to enumerate all solutions.
Due to the geometric nature of these
synthesis problems, classical upper bounds
on the number of solutions, e.g.,
see \cite[Chap.~8]{SoWaBook05},
are often several orders of magnitude
larger than the actual number of solutions.
This paper, inspired by estimation methods
in \cite{CCPC,FRG}, 
develops a method to statistically
estimate the number of solutions using
monodromy loops \cite{Monodromy}.  
The statistical estimates are derived
by viewing a monodromy loop as applying
a capture-mark-recapture model on
a closed population often
used to estimate 
animal~populations~\cite{CaptureRecaptureArticle,CaptureRecaptureBook}.

A shortcoming of previous statistical
estimates is the lack of a 
stopping criterion for showing 
that all solutions have been found.  
This paper overcomes this by incorporating
the multihomogeneous trace test \cite{TraceTest}
for validating~completeness.

The organization of the remainder of
the paper is as follows.
A short summary of related work is provided
in Section~\ref{Sec:background}.
Section~\ref{Sec:StatsEstMonodromy} describes
monodromy and statistical models
for estimating the number of solutions.
Section~\ref{Sec:TraceTest} describes
the trace trace as a stopping criterion.
Section~\ref{Sec:Results} considers
Alt's problem and motion generation
problems for Watt I linkages 
pictorially represented in Figure~\ref{fig:W1and4BarLabeled}(b). 
A short conclusion is provided in Section~\ref{sec:Conclusion}.

\section{Related work}\label{Sec:background}

Since kinematic synthesis problems, such as path synthesis and motion generation~\cite{MechDesign,McCarthyBook},
are classical but challenging problems that 
yield polynomial systems \cite{AlgKinematics}, 
the following 
provides some related work using 
numerical algebraic geometry \cite{Bates2013,SoWaBook05}.  
A variety of homotopy methods have been
used to solve a collection of synthesis problems,
such as \cite{KinSynW1,6BarParallel,WMS9Pt4Bar,HWP3R}.
The estimation method in \cite{FRG} uses
a coupon collector model based on the success
ratio of finding new solutions using
parameter homotopies \cite{Parameter}.
In \cite{CCPC}, the total number of solutions
is estimated using a sequence of 
parameter homotopies between two parameters.

Monodromy \cite{Monodromy} is a standard tool
in numerical algebraic geometry that was first
used to decompose positive-dimensional 
solution components.  
The use of monodromy loops to generate new solutions
to parameterized polynomial systems has been
used in a variety of applications, e.g.,
\cite{MonSolver1, MonSolver2,TensorDecomp}. 
This paper adds statistical estimates based on
using monodromy loops to find 
new~solutions.

The affine trace test \cite{AffineTraceTest}
is also a standard tool in numerical algebraic
geometry that was first used as a stopping
criterion for monodromy when decomposing
positive-dimensional solution components.  
Various versions of the trace test
have been described, such as \cite{TraceTest,TraceTestDeriv,TraceTestLRS}.
This paper uses the multihomogeneous
trace test \cite{TraceTest} for validating
the completeness of the solution set.

\section{Statistical estimation using monodromy loops}\label{Sec:StatsEstMonodromy}

Consider a synthesis problem that is described
by solving a pamareterized polynomial system
$F(x;p) = 0$ consisting of $N$ polynomials 
in the variables $x\in\bC^N$
and parameters $p\in\bC^P$.
The goal is to statistically estimate
the number of isolated solutions $F(x;p^*) = 0$
for generic $p^*$.  Since the set of solutions
to $F(x;p^*) = 0$ is a fixed set, sampling
from the solution set corresponds with sampling
from a closed population.  Section~\ref{SubSec:Monodromy} describes using 
monodromy loops \cite{Monodromy} 
(see also \mbox{\cite[\S~15.4]{SoWaBook05}}) to sample
without replacement in a closed population.
The statistical estimation is based on 
capture-mark-recapture models often used
to estimate animal populations \cite{CaptureRecaptureArticle,CaptureRecaptureBook}.
Section~\ref{SubSec:LPEst}
provides an estimate based on a single trial
using a Lincoln-Petersen estimate
while Section~\ref{SubSec:ChapmanEst} 
provides a Chapman estimate, which is 
an unbiased Lincoln-Petersen estimate.
Section~\ref{SubSec:SchnabelEst} 
describes a Schnabel estimate
based on the results from several trials.

\subsection{Monodromy loops}\label{SubSec:Monodromy}

Given a set $S\subset\bC^N$ 
consisting of distinct isolated 
solutions to $F(x;p^*) = 0$, 
monodromy can be used to generate 
another subset of solutions as follows.
First, one selects a random loop $\gamma\subset\bC^P$
starting and ending at $p^*$, that is, 
\mbox{$\gamma(0) = \gamma(1) = p^*$}.  Then, one utilizes
homotopy continuation (see \cite{Bates2013,SoWaBook05} for a general overview)
to track the solution paths
of $F(x;\gamma(t)) = 0$ from start points~$S$
at $t = 1$ to, say, end points $E$ at $t = 0$.
The set $E$ consists of (possibly different)
solutions to $F(x;p^*)=0$ as 
illustrated in the following.

\begin{example}
Consider $F(x;p) = x^2 - p$ with $p^* = 1$
and $S = \{1\}$.  For the loop $\gamma(t) = e^{2\pi t\sqrt{-1}}$ with $p^*=\gamma(0)=\gamma(1)$, one has $E = \{-1\}$
as shown in Figure~\ref{fig:MonodromyEx}.

\begin{figure}[!ht]
\centering
\begin{picture}(90,90)
\put(-40,-20){\includegraphics[scale=0.3]{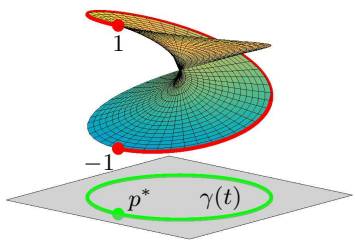}}
\put(28,6){$p^*$}
\put(22,65){$1$}
\put(11,20){$-1$}
\put(55,7){$\gamma(t)$}
\end{picture}
\caption{Illustration of monodromy yielding a new solution}\label{fig:MonodromyEx}
\vspace{-0.6cm}
\end{figure}
\end{example}

Theoretically, 
each solution path of $F(x;\gamma(t)) = 0$
remains on the same irreducible component of 
the solution set of $F(x;p) = 0$ in $\bC^{N}\times\bC^{P}$
and \mbox{$\#E = \#S$} with probability one
for a random loop.  Moreover, such 
solution paths can be 
certifiably tracked, e.g., see \cite{Certification}.  
When using faster heuristic path tracking methods,
failures can occur.  Thus, 
the statistical estimation models (see \cite{CaptureRecaptureArticle,
CaptureRecaptureBook} for more details)
allow for $\#E < \#S$.
In order to find all solutions using monodromy
loops starting with only one solution, 
the monodromy group of the isolated solutions
of $F(x;p) = 0$ must be transitive, e.g., see \cite{GaloisGroup}.  However, this is commonly
the case for synthesis problems, including
the ones in Section~\ref{Sec:Results}.

\subsection{Lincoln-Petersen estimation}\label{SubSec:LPEst}

The first estimate of the total number of solutions
is based on the ratio of repeated solutions
from one monodromy loop.
With start points $S$ and end points $E$,
$S\cap E$ is the set of repeated solutions
while $S\setminus E$ and $E\setminus S$
are the nonrepeated solutions in $S$ and $E$, respectively.
The following is 
the Lincoln-Petersen estimate for the number
of solutions, $\beta$, and variance:
\begin{equation}\label{eq:LPest}
\mbox{\small $
\beta=\dfrac{\#S\cdot \#E}{\#(S\cap E)}
\hbox{\,\,\,\,\,\,and\,\,\,\,\,\,}
\Var(\beta)=\dfrac{(\#S+1)\cdot(\#E+1)\cdot \#(S\setminus E) \cdot \#(E\setminus S)}{(\#(S\cap E)+1)^2\cdot(\#(S\cap E)+2)}.$
}
\end{equation}
Thus, the $95\%$ confidence interval is given by
\begin{equation}\label{eq:ConfidenceInterval}
\mbox{\small $
\left(\beta- 1.96\sqrt{\Var(\beta)},~\beta + 1.96\sqrt{\Var(\beta)}\right).$}
\end{equation}

\subsection{Chapman estimation}\label{SubSec:ChapmanEst}

Since the Lincoln-Petersen estimate
is biased for small sample sizes \cite{Chapman},
the following estimate of Chapman is unbiased:
\begin{equation}\label{eq:Cest}
\mbox{\small $
\beta=\dfrac{(\#S+1)\cdot(\#E+1)}{\#(S\cap E)+1}-1.$
}
\end{equation}
The variance and 95\% confidence interval are given in \eqref{eq:LPest} and \eqref{eq:ConfidenceInterval}, respectively.

\subsection{Schnabel estimation}\label{SubSec:SchnabelEst}

The estimates in \eqref{eq:LPest} and \eqref{eq:Cest}
utilize results from a single monodromy loop.
The Schnabel estimate uses data from several
monodromy loops.  For example, the experimental 
results in Section~\ref{Sec:Results} determine
estimates based on the last 
three monodromy loops, a so-called rolling 
window of size $3$.
For estimating the number of solutions
using data from $\ell\geq 1$ loops, 
suppose that $S^{(k)}$ and $E^{(k)}$
are the start and end points for the $k^{\rm th}$
monodromy loop where $k = 1,\dots,\ell$.  
The Schnabel estimate for the number
of solutions and variance
of the inverse are
\begin{equation}\label{eq:Sest}
\mbox{\small $
\beta=\dfrac{\sum_{k=1}^\ell \#S^{(k)}\cdot \#E^{(k)}}{\sum_{k=1}^\ell \#(S^{(k)}\cap E^{(k)})}$
\,\,\,\,\,and\,\,\,\,\,$
\Var\left(\beta^{-1}\right)=\dfrac{\sum_{k=1}^\ell \#\left(S^{(k)}\cap E^{(k)}\right)}{\left(\sum_{k=1}^\ell \#S^{(k)}\cdot \#E^{(k)}\right)^2}.$
}
\end{equation}
Thus, the $95\%$ confidence interval is given by
\begin{equation}\label{eq:ConfidenceIntervalS}
\mbox{\small $
\left(\left(\beta^{-1} - 1.96\sqrt{\Var(\beta^{-1})}\right)^{-1},~ \left(\beta^{-1} + 1.96\sqrt{\Var(\beta^{-1})}\right)^{-1}\right).$
}
\end{equation}

\section{Trace test}\label{Sec:TraceTest}

One potential indicator that all isolated
solutions have been found is that several random 
monodromy loops fail to yield new solutions,
e.g., as used in \cite{TensorDecomp}.  
The multihomogeneous trace test~\cite{TraceTest}
can be used to validate that every random monodromy
loop will not yield new solutions.  
This confirms all solutions have been found 
when the monodromy group is transitive
(see Section~\ref{SubSec:Monodromy}).

The affine trace test~\cite{AffineTraceTest}
validates completeness if the centroid
of the solutions moves linearly as 
the intersecting linear slice is moved parallelly.
The key to the multihomogeneous trace test~\cite{TraceTest}
is to view $F(x;p)$ as a system in 
$\bC^N\times\bC^P$ with a parallelly 
moving bilinear slice
as shown in the following.

\begin{example}\label{Ex:Trace}
Consider validating that $F(x;p) = (p+1)x^2 - p$
has two roots for $p^* = 1$.  
Thus, one takes a bilinear slice moving
parallelly that contains the linear space $p=p^*$ 
at $t = 0$, say $L_t(x;p) = 4x(p-1)-t$.
For general $t$, $F = L_t = 0$ has~3 solutions
such that $2$ satisfy $p = p^*$ 
and $1$ satisfies $x = 0$ when $t = 0$.
Figure~\ref{fig:TraceTest} plots $F = 0$ (black)
and $L_t = 0$ for $t=0$ (blue),
$t=1$ (red), and $t=-1$ (green)
along with their corresponding centroids lying
on the dashed line $p = 2/3$ (magenta).  
This validates $F(x;p^*)=0$ has $2$ solutions.

%%%%%Picture of trace test
\begin{figure}[!ht]
\centering
%\begin{picture}(120,110)
\begin{picture}(120,128)
%\put(-15,130){\includegraphics[scale=0.25,angle=270]{Trace_pic4.eps}}
\put(-50,-26){\includegraphics[scale=0.29]{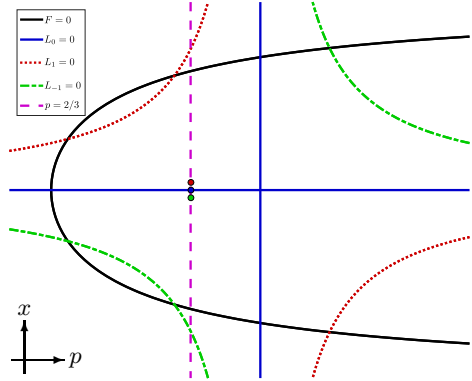}}
%\put(66,-9){$p^*$} %\put(40,-5){$p^*$}
%\put(140,56){$x=0$} %\put(85,55){$x=0$}
%\put(35,120){$2/3$} %\put(22,116){$2/3$}
%\thicklines
  \put(-21,0){\vector(1,0){20}}
  \put(-16,-5){\vector(0,1){20}}
  \put(1,0){\makebox(1,0)[l]{$p$}}
  \put(-16,17){\makebox(0,1)[b]{$x$}}
\end{picture}
\caption{Trace test for $F(x;p) = 0$
using a bilinear slice $L_t(x;p) = 0$ moving
parallelly.}
\label{fig:TraceTest}
\vspace{-0.6cm}
\end{figure}
\end{example}

As illustrated in Example~\ref{Ex:Trace}, 
one disadvantage is that additional solutions
are needed due to the bilinear slice.
Table~\ref{Table:W1Results} compares the number
of solutions to the synthesis problem
and other solutions needed
for validation.

\section{Results}\label{Sec:Results}

The following applies monodromy loops 
using the software {\tt Bertini} \cite{Bertini_Software}
for statistically estimating the number of solutions to several synthesis problems
comparing the Lincoln-Petersen and Chapman
estimates using one loop with the Schnabel estimate
using the last three loops, i.e., a rolling
window of size~$3$.
A comparison of the number of additional
solutions needed to use the trace 
test~is~provided. Data from the 
examples is available at~\url{dx.doi.org/10.7274/r0-qw8q-r924}.

\subsection{Four-bar mechanism}\label{SubSec:4Bar}

\begin{figure*}[!ht]
\begin{center}
\includegraphics[width=1\textwidth]{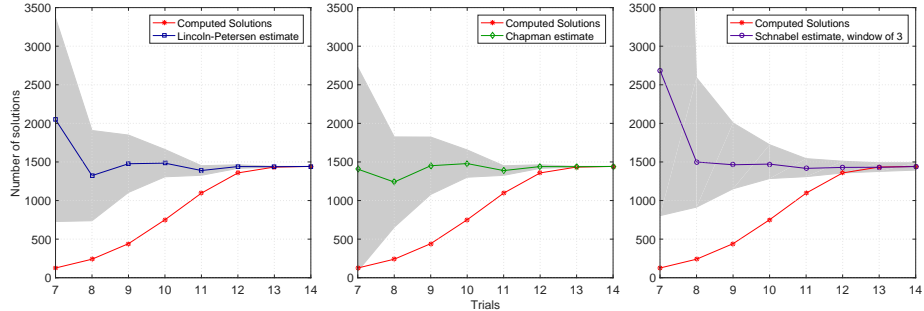}
\end{center}
\vspace{-0.7cm}
\caption{Estimates for solving 
Alt's problem for four-bar linkages as the 
monodromy trials progressed using different estimates
from left to right: Lincoln-Petersen, Chapman, and Schnabel using a window of size 3.
The shaded area is the 95\% confidence interval for each estimate.}
\label{fig:4BarResults}
%\vspace{-0.5cm}
\end{figure*}

Alt's problem~\cite{Alt} for four-bar linkages
is to count 
the number of coupler curves 
that pass through $9$ general points in the plane
(namely, 1442 \cite{WMS9Pt4Bar}).
Figure \ref{fig:4BarResults} shows the number of solutions computed and the various estimation methods as the trials using monodromy loops progressed. 
When there are few repeated solutions, 
the estimate has a large confidence interval
that quickly converges to the actual number of solutions.  In our experiment, by the $10^{\rm th}$ loop in which only $52\%$ of the solutions are 
known, the statistical estimates are 
within 2.9\% of the actual number of solutions.
A trace test validation was performed
in \cite[\S~7.2.1]{TraceTest}.

%%%%%%%%%%%Graph of Four Bar results
%		\begin{figure*}[h]
%		\centering
%\begin{tabular}{ccc}		
%\includegraphics[width=0.29\textwidth]{Bar4_LP.eps}
%& &
%\includegraphics[width=0.29\textwidth]{Bar4_C.eps}
%\\
%(a) Lincoln-Petersen & \hbox{\,\,\,\,} & (b) Chapman \\
%\includegraphics[width=0.29\textwidth]{Bar4_S2.eps}
%& &
%\includegraphics[width=0.29\textwidth]{Bar4_S3.eps} \\
%(c) Schnabel, window of size 2 & \hbox{\,\,\,\,} &
%(d) Schnabel, window of size 3 
%\end{tabular}
%\caption{Estimates for Alt's problem for four-bar linkages as the trials progressed using different estimates. The shaded area is the 95\% confidence interval for each estimate.}
%\label{fig:4BarResults}
%\end{figure*}

\subsection{Watt I six-bar mechanism}\label{SubSec:Watt1}

%%%% Table of Watt 1 results %%%%%%%%%%%%%%%%%%
\begin{table}[!t] \begin{center}
\resizebox{.95\textwidth}{!}{	
\begin{tabular}{|c|c|c|c|c|c|c|}
\hline
$N$ & DOF & fixed pivots &  \# variables& \# solutions & trace test& \# other solutions\\ \hline \hline
6 & 4 & $A,B$ & $50$& 5754 & Yes ($C$) & 7167\\
\hline  
7 & 2 & $A$ & $60$ & 198,614 & Yes ($B$) & 115,126\\
\hline
8 & 0 & -- & $70$ & $\approx 1.68\cdot 10^6$ & -- & --\\
\hline
\end{tabular}
}
\end{center}
\vspace{-0.1cm}
\caption{Table of results for Watt I synthesis
with $N$ task positions} \label{Table:W1Results}
\vspace{-0.7cm}
\end{table}

The last collection of experiments arise from
motion generation problems for Watt I six-bar linkages (Watt IB using the convention in \cite{CogsMadeEasy})
following the formulation in \cite{KinSynW1}.
For~$N$ task positions, there are $16-2N$ degrees
of freedom~(DOF) so it is natural to 
specify one (when $N = 7$) 
or both (when $N = 6$) ground pivots
as additional constraints.
Table \ref{Table:W1Results} summarizes
the setup of the problems and 
results for $N = 6,7,8$ task
positions (see labels in Figure~\ref{fig:W1and4BarLabeled}(a)).  The $N = 6,7$ 
cases were
validated using the trace test applied
to the first task position and the listed
variable.  For $N = 8$, the 
number of solutions is estimated.

\begin{figure*}[!ht]
\begin{center}
\includegraphics[width=1\textwidth]{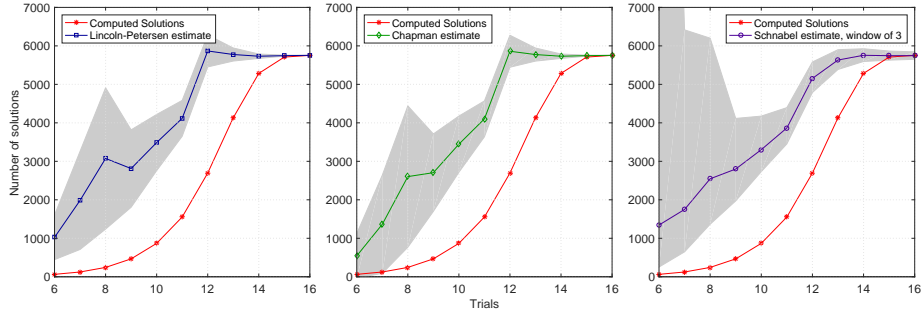}
\end{center}
\vspace{-0.6cm}
\caption{Estimates for Watt I synthesis with
$N = 6$ task positions and both pivots fixed
as the monodromy trials progressed using different estimates from left to right: Lincoln-Petersen, Chapman, and Schnabel using a window of size 3.
The shaded area is the 95\% confidence interval for each estimate.}
\label{fig:W1N6Results}
\vspace{-0.1cm}
\end{figure*}

The $N = 6$ case was considered
in \cite{KinSynW1}, \cite{6BarParallel},
and \cite{CCPC} reporting 5735, 
5743, and 5754 solutions, respectively.
The trace test confirms
the number of solutions is indeed 5754
with results of our monodromy loops and estimations presented in Figure~\ref{fig:W1N6Results}. 
In particular, once monodromy loops
returned over $25\%$ repeats,
the estimations quickly converged to the number of solutions. 

 %%%%%Graph of results of monodromy for W1 N=6 case. 
%\begin{figure*}[h]
%\begin{center}
%\begin{tabular}{ccc}		
%\includegraphics[width=0.29\textwidth]{W1N6LPest.eps}
%& &
%\includegraphics[width=0.29\textwidth]{W1N6Cest.eps}
%\\
%(a) Lincoln-Petersen & \hbox{\,\,\,\,} & (b) Chapman \\
%\includegraphics[width=0.29\textwidth]{W1N6_S2.eps}
%& &
%\includegraphics[width=0.29\textwidth]{W1N6_S3.eps} \\
%(c) Schnabel, window of size 2 & \hbox{\,\,\,\,} &
%(d) Schnabel, window of size 3 
%\end{tabular}
%\caption{Estimates for Watt I synthesis with $N=6$ task positions and both pivots fixed as the trials
%progressed using different estimates
%with 95\% confidence interval.}\label{fig:W1N6Results}
%\end{center}
%\end{figure*}

\begin{figure*}[!ht]
\begin{center}
\includegraphics[width=1\textwidth]{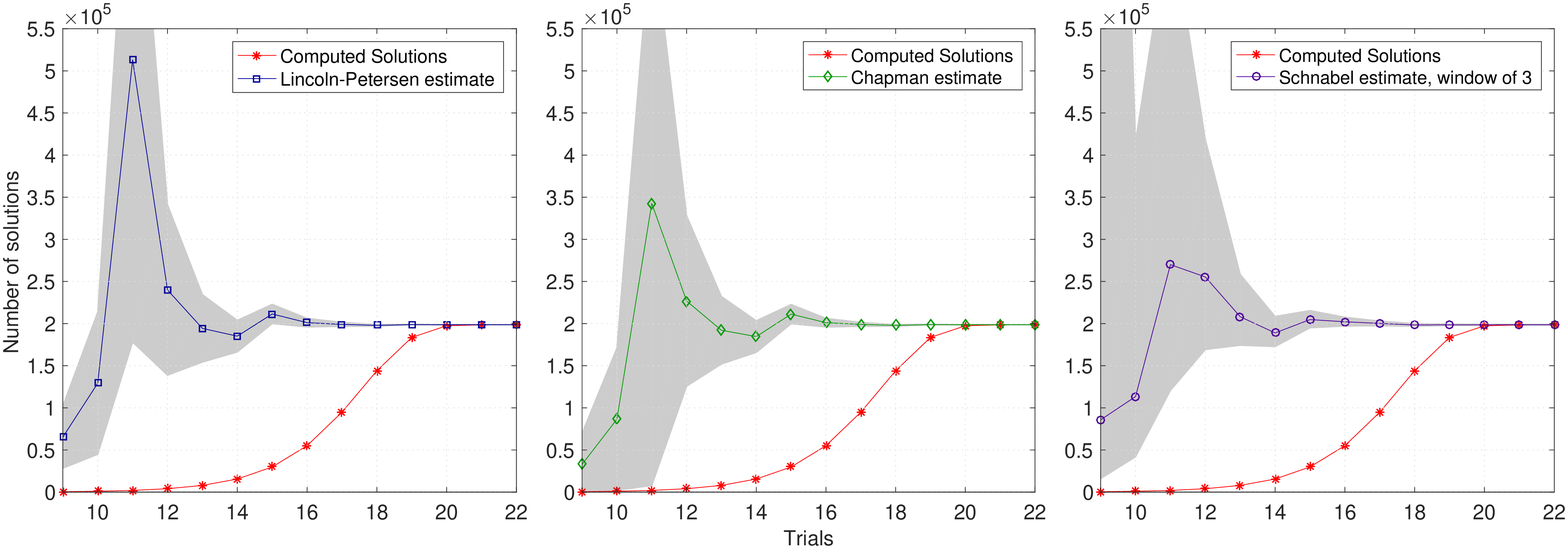}
\end{center}
\vspace{-0.6cm}
\caption{Estimates for Watt I synthesis with
$N = 7$ task positions and pivot $B$ fixed
as the monodromy trials progressed using different estimates from left to right: Lincoln-Petersen, Chapman, and Schnabel using a window of size 3.
The shaded area is the~95\% confidence interval for each estimate.}
\label{fig:W1N7Results}
\vspace{-0.1cm}
\end{figure*}

\begin{figure*}[!ht]
\begin{center}
\includegraphics[width=1\textwidth]{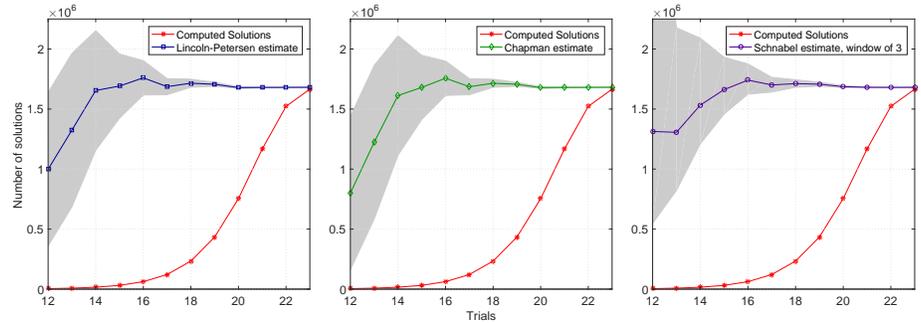}
\end{center}
\vspace{-0.6cm}
\caption{Estimates for Watt I synthesis with
$N = 8$ task positions 
as the monodromy trials progressed using different estimates from left to right: Lincoln-Petersen, Chapman, and Schnabel using a window of size 3.
The shaded area is the~95\% confidence interval for each estimate.}
\label{fig:W1N8results}
\vspace{-0.5cm}
\end{figure*}

The estimations for $N = 7$ with 
pivot $B$ fixed are shown
in Figure~\ref{fig:W1N7Results}
with the trace test validating the
total number of solutions is 198,614.
Once over 50\% of the solutions 
were found, 
the estimate was within 1\% of the actual number. 

An estimate in \cite{CCPC} 
for the fully constrained
motion generation problem with $N = 8$ task positions
is 840,300 cognate pairs, i.e., 1,680,600 distinct mechanisms, which is within the 95\% confidence
intervals of our results presented in
Figure~\ref{fig:W1N8results}. 

\section{Conclusion}\label{sec:Conclusion}

Linkage design in kinematics
naturally leads to solving
parameterized polynomial systems.
A statistical estimation of the
number of solutions to such systems
was developed by using
a capture-mark-recaputure
model based on monodromy loops.
This statistical method
was demonstrated on
Alt's problem and several~Watt I motion generation problems. 
The results show that the estimates
quickly converge to the actual number
of solutions once a reasonable 
proportion of the solutions have been found.
A stopping criterion based
on a multihomogeneous trace test
is used to validate the completeness 
of the solution set.


\bibliographystyle{spmpsci}

\end{document}